%% file: arxiv.tex
\documentclass[11pt]{article}
\usepackage{arxiv}
\usepackage{enumitem}
\usepackage{natbib}
\renewcommand\cite{\citep}          

\usepackage{xcolor}
\definecolor{linkblue}{HTML}{1A4FA0}
\usepackage[colorlinks=true,linkcolor=linkblue,citecolor=linkblue,urlcolor=linkblue]{hyperref}

\input{preamble_shared}

\arxivbuildtrue

\usepackage{mathpazo}
\makeatletter
\renewcommand{\normalsize}{%
  \@setfontsize\normalsize\@xipt{13.6}%
  \abovedisplayskip      8\p@ \@plus 2\p@ \@minus 5\p@
  \abovedisplayshortskip \z@ \@plus 3\p@
  \belowdisplayskip      \abovedisplayskip
  \belowdisplayshortskip 4.5\p@ \@plus 3\p@ \@minus 3\p@
}
\renewcommand{\small}{%
  \@setfontsize\small\@xpt{12}%
  \abovedisplayskip      7\p@ \@plus 1.5\p@ \@minus 4\p@
  \abovedisplayshortskip \z@  \@plus 2\p@
  \belowdisplayskip      \abovedisplayskip
  \belowdisplayshortskip 3.5\p@ \@plus 2\p@ \@minus 2\p@
}
\renewcommand{\footnotesize}{\@setfontsize\footnotesize\@ixpt{11}}
\renewcommand{\scriptsize}{\@setfontsize\scriptsize\@viiipt{9.5}}
\renewcommand{\tiny}{\@setfontsize\tiny\@vipt\@viipt}
\renewcommand{\large}{\@setfontsize\large\@xiipt{14}}
\renewcommand{\Large}{\@setfontsize\Large\@xivpt{18}}
\renewcommand{\LARGE}{\@setfontsize\LARGE\@xviipt{22}}
\makeatother
\normalsize

\makeatletter
\renewenvironment{table*}[1][tbp]{\@float{table}[#1]}{\end@float}
\renewenvironment{figure*}[1][tbp]{\@float{figure}[#1]}{\end@float}
\makeatother

\renewcommand{\colfig}{0.62\linewidth}

\renewcommand{\undertitle}{}
\renewcommand{\headeright}{}

\makeatletter
\renewcommand{\@maketitle}{%
  \vbox{%
    \hsize\textwidth \linewidth\hsize
    \vskip 0.1in
    \centering
    {\LARGE\bfseries \@title\par}
    \vskip 0.28in
    \def\And{\end{tabular}\hfil\linebreak[0]\hfil\begin{tabular}[t]{c}\ignorespaces}%
    \def\AND{\end{tabular}\hfil\linebreak[4]\hfil\begin{tabular}[t]{c}\ignorespaces}%
    \begin{tabular}[t]{c}\@author\end{tabular}%
    \vskip 0.25in
  }%
}
\renewenvironment{abstract}
  {\centerline{\large\bfseries Abstract}\vspace{0.4em}\begin{quote}}
  {\end{quote}}
\makeatother

\title{\benchmark: Can AI Agents Improve Language Models Across Multiple Domains?}

\author{%
  \normalfont
  Haojin Wang$^{1,*}$ \quad
  Yangzhen Wu$^{2,*}$ \quad
  Zeyi Chen$^{2,*}$ \quad
  Jian Xie$^{3}$ \quad
  Jiaying Ye$^{4}$ \\[3pt]
  \normalfont
  Lili Meng$^{5}$ \quad Shuyun Tang$^{2}$ \quad
  Shangding Gu$^{2,\dagger}$ \quad Dawn Song$^{2}$ \\[8pt]
  \normalfont\normalsize
  $^{1}$University of Illinois Urbana-Champaign \qquad
  $^{2}$University of California, Berkeley \\[2pt]
  \normalfont\normalsize
  $^{3}$The Ohio State University \qquad
  $^{4}$University of Washington \qquad $^{5}$Google
}

\date{\vspace{-0.32in}}

\makeatletter
\newcommand\blfootnote[1]{%
  \begingroup\renewcommand\thefootnote{}\footnote{#1}%
  \addtocounter{footnote}{-1}\endgroup}
\makeatother

\begin{document}
\maketitle
\blfootnote{$^{*}$Equal contribution. $^{\dagger}$Corresponding author: \href{mailto:shangding.gu@berkeley.edu}{\texttt{shangding.gu@berkeley.edu}}.}

\input{sections/fig_overview}
\input{sections/0_abstract}
\input{sections/1_introduction}
\input{sections/2_setup}
\input{sections/3_experiment}

\input{sections/4_analysis}
\input{sections/5_related_work}
\input{sections/6_conclusion}
\input{sections/7_limitation}

\bibliographystyle{acl_natbib}
\bibliography{reference}

\clearpage
\appendix
\input{sections/appendix}

\end{document}

%% file: preamble_shared.tex
\providecommand{\colfig}{\linewidth}
\newif\ifarxivbuild \arxivbuildfalse
\usepackage{times}
\usepackage{latexsym}
\usepackage[T1]{fontenc}
\usepackage[utf8]{inputenc}
\usepackage{microtype}
\usepackage{inconsolata}

\usepackage{amsmath,amssymb}
\usepackage{graphicx}
\usepackage{xcolor}
\usepackage{xspace}
\usepackage{url}
\usepackage{booktabs}
\usepackage{array}
\usepackage{tabularx}
\usepackage{makecell}
\usepackage{multirow}
\usepackage{colortbl}
\usepackage{pifont}
\usepackage{rotating}
\usepackage{wrapfig}
\usepackage{algorithm}
\usepackage{algpseudocode}
\usepackage{caption}
\usepackage{tcolorbox}
\tcbuselibrary{breakable}
\usepackage{fvextra}
\usepackage{cleveref}
\definecolor{takeawaybg}{HTML}{EAF2FC}
\definecolor{takeawayrule}{HTML}{9EC5F4}
\definecolor{promptgrey}{HTML}{8A8A82}
\definecolor{promptorange}{HTML}{D97757}
\definecolor{promptblue}{HTML}{4C6EF5}
\definecolor{promptgreen}{HTML}{2F9E44}
\definecolor{promptpurple}{HTML}{9775FA}
\newtcolorbox{promptbox}[2][promptgrey]{breakable, colback=#1!5, colframe=#1!55, boxrule=0.6pt, arc=2pt,
  left=5pt, right=5pt, top=4pt, bottom=4pt, title={\small\bfseries #2}, coltitle=black,
  colbacktitle=#1!22}
\newtcolorbox{takeaway}{colback=takeawaybg, colframe=takeawayrule, boxrule=0.6pt, arc=2pt,
  left=6pt, right=6pt, top=4pt, bottom=4pt, before skip=8pt, after skip=8pt}

\newtcolorbox{notebox}[1]{breakable, colback=promptgreen!4, colframe=promptgreen!55,
  boxrule=0.6pt, arc=2pt, left=5pt, right=5pt, top=4pt, bottom=4pt, coltitle=black,
  colbacktitle=promptgreen!22, title={\small\bfseries #1}}
\newcommand{\noteline}[3]{\par\addvspace{3pt}\noindent
  {\ttfamily\scriptsize\textbf{[#1 \textcolor{promptgreen!55!black}{| #2}]} #3\par}}
\newcommand{\roundline}[3]{\par\addvspace{4pt}\noindent
  \colorbox{promptgreen!13}{\parbox{\dimexpr\linewidth-2\fboxsep\relax}
  {\scriptsize\textbf{#1}~ #2\par\addvspace{1pt}\textit{#3}}}\par\addvspace{2pt}}

\newtcolorbox{roundbox}[1]{breakable, colback=promptblue!3, colframe=promptblue!50, boxrule=0.6pt,
  arc=2pt, left=5pt, right=5pt, top=4pt, bottom=4pt, title={\small\bfseries #1}, coltitle=black,
  colbacktitle=promptblue!18}
\newcommand{\stepline}[3]{\par\addvspace{3.5pt}\noindent
  {\scriptsize\textbf{#1.~#2}\par\addvspace{0.5pt}\textcolor{black!65}{#3}\par}}

\newcommand{\cmark}{\ding{51}}
\newcommand{\xmark}{\ding{55}}

\newcommand{\benchmark}{\textsc{ParetoRSI-Gym}\xspace}

\colorlet{deltapos}{teal!60!black}
\colorlet{deltaneg}{red!70!black}

\crefname{section}{\S\!}{\S\S\!}
\crefname{table}{Tab.}{Tabs.}
\crefname{figure}{Fig.}{Figs.}
\floatname{algorithm}{Loop}
\crefname{appendix}{App.}{Apps.}
\crefname{equation}{Eq.}{Eqs.}

%% file: sections/fig_overview.tex
\ifarxivbuild\begin{figure*}[!ht]\else\begin{figure*}[t]\fi
        \centering
        \includegraphics[width=\textwidth]{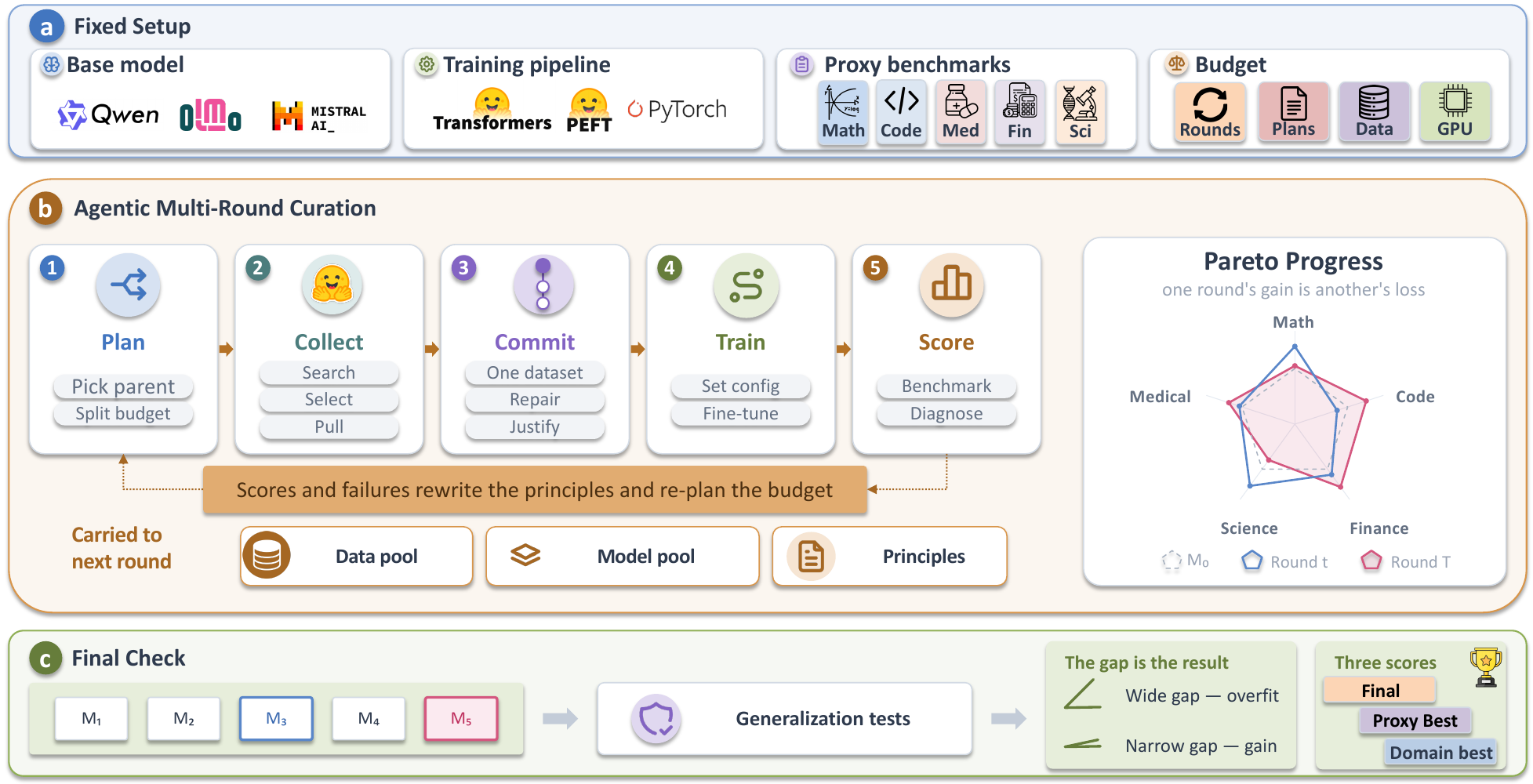}
        \caption{\textbf{Overview of \benchmark}. 
        Starting from a fixed initial model, an agent curates post-training data for five domains over repeated rounds. 
        In each round the current principles direct which datasets the agent searches for and pulls from HuggingFace, and which entries it keeps; the resulting entries are committed to a persistent pool, which is submitted for fine-tuning and tested on the per-domain benchmarks; the scores are then used to update the principles for the next round. 
        Success requires the capability vector to expand on every domain at once rather than on a single target.}
        \label{fig:overview}
        \end{figure*}

%% file: sections/0_abstract.tex
\begin{abstract}
Recursive self-improvement (RSI) requires an agentic system to design, train, and evaluate its own successor.
Data-driven post-training is a central part of that loop, yet existing evaluations measure it on a single benchmark, where an agent selects from a static pool or distills a stronger model to push one score up in one run.
Real post-training is multi-domain, and a gain in one domain has to be weighed against a regression in another.
We propose \benchmark, in which an agent searches for data, curates it, trains, reads back per-domain results, and repeats, with everything but the data held fixed: the initial LLM, the five domains (math, code, finance, medicine, science), the toolset, and the SFT pipeline.
Generalization benchmarks in each domain separate capability gains from overfitting to the scores the agent optimizes against.
Across three agent systems, three base models, five rounds and multiple seeds, the strongest system, Claude Code with \texttt{Opus-5}, raises the generalization scores of its proxy-best checkpoint by $9.2\%$ relative to the starting model ($+2.3$ absolute points), averaged with equal weight over three base models and five domains, with a single-domain relative gain of up to $63\%$ (medicine on Ministral-3-3B).
Further analysis shows that the harness matters as much as the model it drives, and that each agent keeps a unique data selection strategy.
We see our work as a step towards evaluation of RSI systems in real-world machine learning engineering.\ifarxivbuild\footnote{Code: \url{https://github.com/haojinw0027/ParetoRSI-GYM}.}\fi
\end{abstract}

%% file: sections/1_introduction.tex
\section{Introduction}
Recursive Self-Improvement (RSI) has become a popular research direction in modern agentic systems, where systems are designed to improve themselves by iteratively refining their internal state~\citep{novikov2025alphaevolve, zhang2026darwin}.
A central pathway toward recursive self-improvement is the automation of AI research and development: systems that can propose, implement, and evaluate improvements to subsequent AI systems~\citep{schmidhuber2007godel,chen2026recursive,zhang2026hyperagents}. 
Data-centric post-training is one important component of this loop.
With the advancement of AI agents, recent works have started to explore the automation of the curation process of post-training data~\citep{rank2026posttrainbench,kang2026can,meng2026rsibench}.
For instance, PostTrainBench~\citep{rank2026posttrainbench} explores how well LLM agents can perform post-training autonomously under bounded compute constraints.
Curation-Bench~\citep{kang2026can} evaluates the ability of LLM agents to curate a clean dataset from a given pool of data.
RSIBench-Data~\citep{meng2026rsibench} investigates whether LLM agents can iteratively improve a dataset by adding new data and removing incorrect entries given a target benchmark.
BenchEvolver~\citep{wu2026benchevolver} instead synthesizes data: it evolves existing coding problems into harder, verifiable tasks that serve both as evaluation and as training signal for self-improvement.
In practice, post-training targets a set of capabilities at once: released models are trained on mixtures spanning mathematics, code, science, and professional domains, and are reported per capability rather than on a single score~\citep{lambert2024tulu,yang2025qwen3,wang2025nemotron}.
Thus, instead of evaluating the capability of modern agents in post-training data curation for a single benchmark score, in this work, we ask: \textbf{can LLM agents curate post-training data that improves a model across multiple capabilities at once, and keep improving it over repeated rounds of curation?}

\begin{table*}[t]
    \centering
    \scriptsize
    \ifarxivbuild\setlength{\tabcolsep}{5pt}\else\setlength{\tabcolsep}{10pt}\fi
    \begin{tabular}{@{}lccccc@{}}
    \toprule
    & \shortstack{Cross-Round \\ State} & \shortstack{Dynamic \\ Data Sources} & \shortstack{Feedback-driven \\ Revision} & \shortstack{Multi-Domain \\ Objective} & \shortstack{Generalization \\ Evaluation} \\
    \midrule
    PostTrainBench~\citep{rank2026posttrainbench} & \cmark & \cmark & \cmark & \xmark & \xmark \\
    Agent$^2$ RL-Bench~\citep{chen2026agent} & \xmark & \xmark & \xmark & \xmark & \xmark \\
    Curation-Bench~\citep{kang2026can} & \cmark & \xmark & \cmark & \xmark & \xmark \\
    DataPrep-Bench~\citep{liang2026dataprep} & \xmark & \xmark & \xmark & \xmark & \xmark \\
    RSIBench-Data~\citep{meng2026rsibench} & \cmark & \cmark & \cmark & \xmark & \xmark \\
    \midrule
    \textbf{\benchmark} & \cmark & \cmark & \cmark & \cmark & \cmark \\
    \bottomrule
    \end{tabular}
    \caption{Comparison with existing benchmarks. \benchmark enforces a more realistic setting for automation of post-training data curation compared to existing benchmarks. A \cmark{} denotes a property enforced by the protocol.}
    \label{tab:positioning}
    \end{table*}

To evaluate the current capability of LLM agents in post-training data curation in a more realistic setting, we propose \benchmark, a new benchmark for post-training data curation that evaluates the ability of LLM agents to automatically curate post-training data for different sets of capabilities, and continuously evolve during the post-training process.
\benchmark begins with a loop-like design, where agents are required to curate data for different domains at once, and then design training recipes based on the curated data.
We define five downstream domains for agents to curate data for: mathematics, code, science, medical and finance.
Then the agents are required to evaluate the finetuned model on the proxy benchmarks for each domain, and then take such feedback to iteratively refine the data curation process and the training recipes.
The agents are allowed to search, retrieve and select data from Hugging Face, and we fix the training recipes as LoRA SFT~\citep{hu2021lora} and let the agents design specific training hyperparameters for each round.
To mitigate data contamination and overfitting to the proxy benchmarks which we use as the feedback during the curation process, we explicitly develop rejection criteria to reject data that is close to the evaluation data, and design a held-out generalization benchmark to evaluate the performance of the agents on same domains but different benchmarks.
This differentiates \benchmark from existing benchmarks such as PostTrainBench~\citep{rank2026posttrainbench} and RSI-Bench Data~\citep{meng2026rsibench} which evaluate the ability of agents to perform post-training on the same benchmark.
We provide a detailed comparison with existing benchmarks in \cref{tab:positioning}.

We evaluate three agent systems, each combining a different harness, on three target models drawn from different pretraining families (3B--7B), for nine independent curation runs as main experiments.
For each run we report two checkpoints: the model from the final round, and the model from the round whose proxy score is highest.
To further understand the behavior of the agents during the curation process, we also report the model from the round whose proxy score is highest.
Claude Code with \texttt{Opus-5} does best: at its proxy-best checkpoint, its generalization scores are $9.2\%$ higher than the starting model's in relative terms ($+2.3$ absolute points), averaged with equal weight over the three base models and five domains.
Gains concentrate in medical and finance, where the largest increase at the proxy-best round is $14.8$ absolute points.
We also observe that the agent systems tend to peak before they stop: in seven of the nine configurations the proxy-best round is not the last one, indicating current agents lack the capability to robustly continue improving the model.
We further carry out a series of analysis to understand the behavior of the agents during the curation process.
With the backbone held fixed and only the harness varied, the outcome still ranges from $+4.3$ points to $+0.2$, highlighting the importance of the harness in the autonomous curation process.
We also conduct a series of analysis on the dataset different agents select. 
We find that two runs of the same agent overlap at $0.24$ Jaccard on the datasets they select, against $0.07$ for different agents, indicating that each agent keeps unique selection strategies.

Overall, our contributions are:

\begin{itemize}[leftmargin=*]
    \item \textbf{Multi-domain RSI benchmark.}
    We introduce \textsc{ParetoRSI-Gym}, a benchmark for iterative, feedback-driven post-training data curation across five domains under bounded resource budgets.

    \item \textbf{Generalization-aware evaluation.}
    We separate proxy feedback from held-out evaluation and screen training data for evaluation overlap to assess generalization, cross-domain trade-offs, and regressions across rounds.

    \item \textbf{Systematic empirical analysis.}
    Experiments across three agent systems and three base models reveal transferable gains, sensitivity to harness design, later-round regressions, and persistent agent-specific data preferences.
\end{itemize}


%% file: sections/2_setup.tex
\section{\benchmark}
\label{sec:setup}

\ifarxivbuild\else\input{sections/fig_overview}\fi

\subsection{Design Principles}
\benchmark is built around five properties that reflect the real-world post-training engineering scenario (\cref{tab:positioning}), and we describe each of them in the following.
\textbf{Cross-round state.} The model pool, the data pool, and the written principles persist across rounds, so what the agent commits in round $t$ constrains round $t{+}1$.
\textbf{Dynamic data sources.} Entries are searched for and pulled from the HuggingFace Hub at run time rather than drawn from a pre-assembled pool.
\textbf{Feedback-driven revision.} Per-domain proxy results, down to individual generations, are returned every round, and the agent must revise its principles on that basis.
\textbf{Multi-domain objective.} The agent curates for $N$ domains at once, sees the full per-domain score vector every round, and is asked to advance the Pareto frontier of that vector rather than any single coordinate.
\textbf{Generalization evaluation.} Final scores come from generalization benchmarks the agent never observes, which separates genuine capability gains from overfitting to the feedback it optimizes against to mitigate the risk of reward hacking~\citep{wang2026rethinking}.

\subsection{Setup}
\cref{fig:overview} gives an overview of \benchmark.
Consider a loop-like curation process for $N$ domains started from an initial model $M_0$.
We first define a model pool $\mathcal{M} = \{M_0\}$ where the agent can store the models it has curated.
The agent starts from $\mathcal{M} = \{M_0\}$, submits to a training pipeline, and is scored by a fixed evaluation protocol.
Evaluation has two parts. 
A proxy benchmark set $\mathcal{T}_{\text{p}} = \{\mathcal{T}^1_{\text{p}},\dots,\mathcal{T}^N_{\text{p}}\}$ reports one score per curated domain and is returned to the agent every round. 
A generalization benchmark set $\mathcal{T}_{\text{g}} = \{\mathcal{T}^1_{\text{g}},\dots,\mathcal{T}^N_{\text{g}}\}$ covers the capability the agent has been trained on but is never exposed during the run. 
The objective of the agent is to advance the attainable frontier of $\mathcal{T}_{\text{p}}(M_t) \in \mathbb{R}^N$ relative to $\mathcal{T}_{\text{p}}(M_0)$ rather than to raise any single coordinate.
Every round the agent receives the full vector $\mathbf{s}_t$ for each model in $\mathcal{M}$ together with the set of non-dominated checkpoints, so it never steers on a single aggregate score.
It carries three objects across rounds: the model pool $\mathcal{M}$, the training pool $P$, whose entries record their source and a stated reason for their commit, and a set of written principles $R$ under a fixed budget $\beta$ for each round.
Because $R$ is the only channel through which evidence from one round reaches the next, its budget bounds how much the agent can accumulate.

\subsection{Data Curation Loop}
We develop a loop-like curation system that allows the agent to improve the model capability in a realistic post-training engineering scenario.
To begin with, the agent is given $\mathcal{M} = \{M_0\}$ and its initial capability vector $\mathbf{s}_0 = \mathcal{T}_{\text{p}}(M_0)$. 
The agent then writes an initial set of principles as $R_0$ based on the initial capability vector $\mathbf{s}_0$ and the budget $\beta_0$.
After that, it starts to curate the post-training data pool $P$ for each domain based on the data principle $R_0$ and the budget $\beta_0$.
All data come from the HuggingFace Hub, and the agent reaches it through three operations.
\textsc{Search} queries the Hub by keywords, tags, and dataset metadata, and returns candidate datasets with their cards.
\textsc{Select} picks which of these datasets, and which splits or subsets of them, are worth acquiring.
\textsc{Pull} downloads the selected data so the agent can inspect the entries and add those it keeps to the pool.
The agent decides what to search for and what to pull based on the data principle $R_0$, within the budget $\beta_0$.

After the first round data pool $P$ is curated, the agent fine-tunes the base model $M_0$ on $P$ through the training pipeline $\mathcal{S}$ along with a training config $\mathcal{C}$ to get the first round model $M_1$, and adds $M_1$ to the model pool $\mathcal{M}$.
The agent then evaluates the first round model $M_1$ on the per-domain benchmarks $\mathcal{T}_{\text{p}}$ to get the first round capability vector $\mathbf{s}_1 = \mathcal{T}_{\text{p}}(M_1)$.
After the first round, the agent first decides which model to use for the next round based on current model pool $\mathcal{M}$.
The agent then updates the principles for the next round according to the proxy evaluation, updating data pool $P$ and training the chosen model $M_{t-1}$ from $\mathcal{M}$ on $P$. 
The loop continues for $T$ rounds.
We finally report the performance of two checkpoints on the generalization benchmarks.
We first report the final round model as the final evaluation result.
We also report the model with best proxy score as the proxy-best evaluation result.
A detailed description of the loop is provided in \cref{app:loop}.

\subsection{Evaluation}
The evaluation of \benchmark is conducted in two parts.
The first part is the proxy evaluation, which is used during training to guide the agent to curate data.
The second part is the generalization evaluation which is distinct from the proxy evaluation to avoid overfitting or data contamination.
\cref{tab:domains} lists both suites: each pair targets the same capability but shares no item, and for code the held-out suite is a separate benchmark.
We report accuracy on the generalization benchmarks.

Because the reported score comes from $\mathcal{T}_{\text{g}}$, full access to $\mathcal{T}_{\text{p}}$ does not expose the items we report on, so we return the whole evaluation trace rather than one number per domain.
The agent receives not only the proxy scores but also the evaluation trace of the model on the proxy benchmarks.
It can therefore attribute a failure to a missing skill or to a format mismatch, and rewrite its principles against that evidence instead of steering on five scalars.
The generalization suite is never exposed during the run and is scored only after the final round, under the same decoding protocol.
A model that gains on the proxy without gaining there has fitted the feedback rather than the capability, and we report the two scores separately.

%% file: sections/3_experiment.tex
\newcommand{\lgo}[2][1.0em]{\makebox[1.6em][l]{\raisebox{-0.28\height}{\includegraphics[height=#1]{figs/logos/#2}}}}
\newcommand{\cclogo}{\makebox[1.6em][l]{\raisebox{-0.28\height}{\includegraphics[trim=580 180 0 0,clip,height=1.0em]{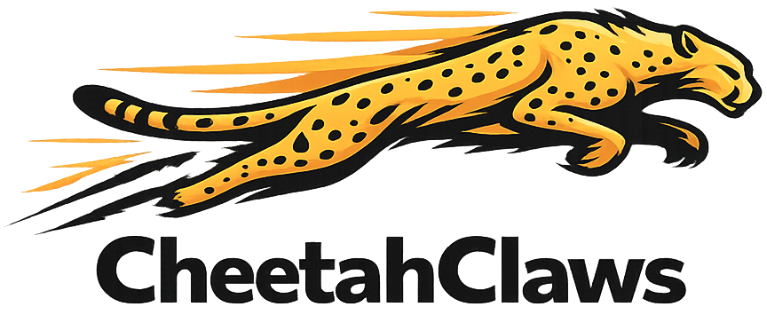}}}}
\newcommand{\mlgo}[2][1.25em]{\makebox[1.85em][l]{\raisebox{-0.28\height}{\includegraphics[height=#1]{figs/logos/#2}}}}
\newcommand{\nolgo}{\makebox[1.6em][l]{}}
\newcommand{\mnolgo}{\makebox[1.85em][l]{}}
\newcommand{\setmzero}[5]{\gdef\mzA{#1}\gdef\mzB{#2}\gdef\mzC{#3}\gdef\mzD{#4}\gdef\mzE{#5}}
\newcommand{\fmtone}[1]{\fpeval{round(#1,1)}\ifnum\fpeval{round(#1,1)==trunc(round(#1,1))}=1\relax.0\fi}
\newcommand{\rc}[2]{#2\%\edef\rcv{\fpeval{round((#2-#1)/#1*100,1)}}%
  \ifcase\fpeval{sign(\rcv)+1}\relax
    \textcolor{deltaneg}{\scriptsize($-$\fmtone{abs(\rcv)}\%)}%
  \or\textcolor{gray}{\scriptsize(0.0\%)}%
  \or\textcolor{deltapos}{\scriptsize(+\fmtone{\rcv}\%)}%
  \fi}

\definecolor{mzerobg}{gray}{0.90}
\newcommand{\dlfmt}[1]{\edef\dlv{\fpeval{round(#1,1)}}%
  \ifdim\dlv pt>0pt\relax\textcolor{deltapos}{$+$\fmtone{\dlv}}%
  \else\ifdim\dlv pt<0pt\relax\textcolor{deltaneg}{$-$\fmtone{abs(\dlv)}}%
  \else\textcolor{gray}{0.0}\fi\fi}
\newcommand{\pc}[2]{#2\,{\scriptsize\dlfmt{#2-\csname mz#1\endcsname}}}
\newcommand{\ms}[3]{\pc{#1}{#2}}

\newcommand{\mz}[1]{\cellcolor{mzerobg}#1}

\section{Experiments}
\subsection{Model}
To start the experiments, we use three different base models and one instruction-tuned model, ranging from different families of models: Qwen3-4B-Base~\citep{yang2025qwen3}, Ministral-3-3B-Base-2512~\citep{liu2026ministral}, and Olmo-3-1025-7B~\citep{olmo2025olmo}.
Such design choice reflects the real-world practice of post-training, where the base model is first pre-trained on large scale data, and then finetuned on domain-specific data by downstream users.
We intentionally choose different base models from different families to mitigate the effect of model family bias.
We also include models with various sizes to measure whether the agent systems can identify different patterns of improvement across different model sizes.

\begin{table}[t]
  \centering
  \footnotesize
  \setlength{\tabcolsep}{3pt}
  \resizebox{\colfig}{!}{%
  \begin{tabular}{@{}l l l@{}}
  \toprule
  \textbf{Domain} & \shortstack[l]{\textbf{Proxy}\\ \textbf{Evaluation}} & \shortstack[l]{\textbf{Generalization}\\ \textbf{Evaluation}} \\
  \midrule
  Math  & \shortstack[l]{MATH500 ID\\ \citep{hendrycks2021measuring}} & \shortstack[l]{MATH500 OOD\\ \citep{hendrycks2021measuring}} \\
  \addlinespace[4pt]
  Science & \shortstack[l]{GPQA ID\\ \citep{rein2023gpqa}} & \shortstack[l]{GPQA OOD\\ \citep{rein2023gpqa}} \\
  \addlinespace[4pt]
  Medical & \shortstack[l]{MedBullets ID\\ \citep{chen2025benchmarking}} & \shortstack[l]{MedBullets OOD\\ \citep{chen2025benchmarking}} \\
  \addlinespace[4pt]
  Finance & \shortstack[l]{SuperGPQA Finance ID\\ \citep{du2025supergpqa}} & \shortstack[l]{SuperGPQA Finance OOD\\ \citep{du2025supergpqa}} \\
  \addlinespace[4pt]
  Code    & \shortstack[l]{MBPP+\\ \citep{austin2021program}} &  
  \shortstack[l]{HumanEval +\\ \citep{chen2021evaluating}} \\
  \bottomrule
  \end{tabular}}
  \caption{The five domains in \benchmark, accompanied by corresponding proxy benchmarks and generalization benchmarks.
  Proxy benchmarks are used within the loop to guide the agent to curate data, and generalization benchmarks are used to evaluate the agent's capability after the loop.}
  \label{tab:domains}
  \end{table}

\begin{table*}[!t]
\centering
\small
\renewcommand{\arraystretch}{0.88}%
\setlength{\aboverulesep}{0.15ex}\setlength{\belowrulesep}{0.2ex}
\setlength{\tabcolsep}{2pt}
\resizebox{\textwidth}{!}{%
\begin{tabular}{@{}l l l ccccc@{}}
\toprule
\textbf{Base Model} & \textbf{Agent} & \textbf{Checkpoint} & \textbf{Math} & \textbf{Science} & \textbf{Medical} & \textbf{Finance} & \textbf{Code} \\
\midrule
\setmzero{65.33}{31.94}{39.36}{24.50}{65.03}%
\multirow{10}{*}[0.5em]{\mlgo[1.25em]{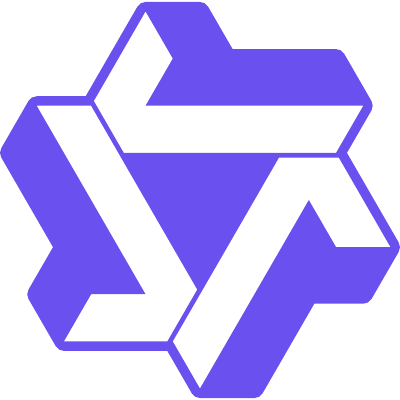}Qwen3-4B-Base}
& \mz{\nolgo$M_0$ (no curation)} & \mz{--} & \mz{65.33} & \mz{31.94} & \mz{39.36} & \mz{24.50} & \mz{65.03} \\
\cmidrule(l{2pt}){2-8}
& \lgo[1.0em]{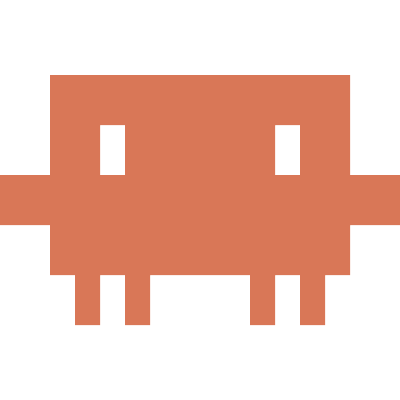}Claude Code & Final & \ms{A}{58.67}{5.62} & \ms{B}{30.13}{1.05} & \ms{C}{40.76}{1.03} & \ms{D}{22.67}{2.84} & \ms{E}{70.50}{1.68} \\
& \nolgo \texttt{Opus-5} & Proxy Best & \ms{A}{60.97}{1.70} & \ms{B}{33.33}{2.67} & \ms{C}{43.69}{2.73} & \ms{D}{24.50}{0.87} & \ms{E}{71.93}{1.96} \\
& \nolgo & Domain Best & \ms{A}{62.08}{1.23} & \ms{B}{35.35}{1.01} & \ms{C}{44.37}{1.56} & \ms{D}{26.17}{1.89} & \ms{E}{72.09}{1.77} \\
\cmidrule(l{2pt}){2-8}
& \lgo[1.0em]{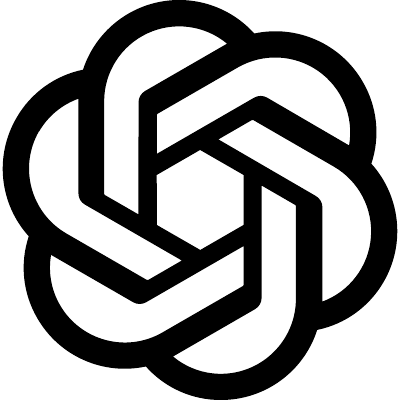}Codex & Final & \ms{A}{65.11}{0.85} & \ms{B}{31.57}{1.53} & \ms{C}{42.12}{0.71} & \ms{D}{24.91}{1.83} & \ms{E}{63.86}{6.88} \\
& \nolgo \texttt{GPT-5.6 Sol} & Proxy Best & \ms{A}{65.11}{0.85} & \ms{B}{31.57}{1.53} & \ms{C}{42.12}{0.71} & \ms{D}{24.91}{1.83} & \ms{E}{63.86}{6.88} \\
& \nolgo & Domain Best & \ms{A}{67.61}{1.59} & \ms{B}{33.71}{1.16} & \ms{C}{43.52}{1.53} & \ms{D}{27.04}{1.01} & \ms{E}{70.35}{1.24} \\
\cmidrule(l{2pt}){2-8}
& \cclogo CheetahClaw & Final & \pc{A}{56.67} & \pc{B}{30.43} & \pc{C}{44.09} & \pc{D}{28.63} & \pc{E}{66.41} \\
& \nolgo \texttt{ds-v4-flash} & Proxy Best & \pc{A}{56.67} & \pc{B}{30.43} & \pc{C}{44.09} & \pc{D}{28.63} & \pc{E}{66.41} \\
& \nolgo & Domain Best & \pc{A}{56.75} & \pc{B}{33.21} & \pc{C}{46.11} & \pc{D}{29.13} & \pc{E}{67.02} \\
\midrule
\setmzero{38.92}{27.65}{23.65}{17.62}{26.69}%
\multirow{10}{*}[0.5em]{\makecell[l]{\mlgo[1.25em]{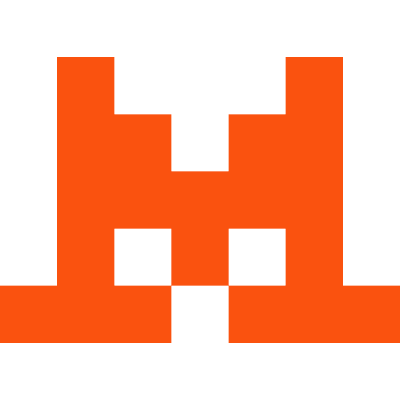}Ministral-3-3B\\ \mnolgo Base-2512}}
& \mz{\nolgo$M_0$ (no curation)} & \mz{--} & \mz{38.92} & \mz{27.65} & \mz{23.65} & \mz{17.62} & \mz{26.69} \\
\cmidrule(l{2pt}){2-8}
& \lgo[1.0em]{claudecode.pdf}Claude Code & Final & \ms{A}{34.00}{3.04} & \ms{B}{28.03}{2.22} & \ms{C}{36.77}{1.86} & \ms{D}{19.87}{0.82} & \ms{E}{38.75}{0.77} \\
& \nolgo \texttt{Opus-5} & Proxy Best & \ms{A}{34.64}{5.32} & \ms{B}{29.00}{1.42} & \ms{C}{38.46}{2.71} & \ms{D}{19.63}{0.65} & \ms{E}{39.26}{1.62} \\
& \nolgo & Domain Best & \ms{A}{36.67}{5.33} & \ms{B}{30.10}{0.51} & \ms{C}{38.96}{1.96} & \ms{D}{20.12}{0.69} & \ms{E}{40.44}{1.15} \\
\cmidrule(l{2pt}){2-8}
& \lgo[1.0em]{openai.pdf}Codex & Final & \ms{A}{36.14}{2.55} & \ms{B}{24.96}{0.86} & \ms{C}{30.57}{1.95} & \ms{D}{17.96}{1.58} & \ms{E}{30.37}{4.64} \\
& \nolgo \texttt{GPT-5.6 Sol} & Proxy Best & \ms{A}{36.58}{2.84} & \ms{B}{26.05}{0.62} & \ms{C}{30.86}{5.50} & \ms{D}{18.58}{1.19} & \ms{E}{28.68}{3.15} \\
& \nolgo & Domain Best & \ms{A}{36.58}{2.84} & \ms{B}{28.03}{2.21} & \ms{C}{35.70}{1.90} & \ms{D}{20.21}{0.40} & \ms{E}{34.97}{0.61} \\
\cmidrule(l{2pt}){2-8}
& \cclogo CheetahClaw & Final & \pc{A}{35.17} & \pc{B}{27.40} & \pc{C}{30.41} & \pc{D}{19.25} & \pc{E}{27.61} \\
& \nolgo \texttt{ds-v4-flash} & Proxy Best & \pc{A}{36.00} & \pc{B}{29.29} & \pc{C}{33.45} & \pc{D}{17.62} & \pc{E}{29.75} \\
& \nolgo & Domain Best & \pc{A}{36.75} & \pc{B}{31.19} & \pc{C}{34.97} & \pc{D}{19.25} & \pc{E}{32.36} \\
\midrule
\setmzero{71.08}{22.73}{32.43}{18.25}{48.01}%
\multirow{10}{*}[0.5em]{\mlgo[1.05em]{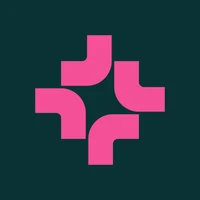}Olmo-3-1025-7B}
& \mz{\nolgo$M_0$ (no curation)} & \mz{--} & \mz{71.08} & \mz{22.73} & \mz{32.43} & \mz{18.25} & \mz{48.01} \\
\cmidrule(l{2pt}){2-8}
& \lgo[1.0em]{claudecode.pdf}Claude Code & Final & \ms{A}{68.72}{2.29} & \ms{B}{18.65}{2.22} & \ms{C}{33.05}{0.80} & \ms{D}{20.42}{2.45} & \ms{E}{46.78}{4.22} \\
& \nolgo \texttt{Opus-5} & Proxy Best & \ms{A}{68.16}{12.80} & \ms{B}{20.71}{5.43} & \ms{C}{32.82}{1.95} & \ms{D}{20.67}{5.20} & \ms{E}{49.44}{0.90} \\
& \nolgo & Domain Best & \ms{A}{72.61}{6.72} & \ms{B}{27.65}{0.33} & \ms{C}{35.41}{2.69} & \ms{D}{22.17}{3.80} & \ms{E}{50.05}{1.95} \\
\cmidrule(l{2pt}){2-8}
& \lgo[1.0em]{openai.pdf}Codex & Final & \ms{A}{66.75}{7.18} & \ms{B}{25.72}{4.39} & \ms{C}{29.62}{1.03} & \ms{D}{17.96}{1.51} & \ms{E}{49.59}{0.73} \\
& \nolgo \texttt{GPT-5.6 Sol} & Proxy Best & \ms{A}{69.50}{3.36} & \ms{B}{24.92}{3.66} & \ms{C}{30.41}{0.34} & \ms{D}{19.00}{1.44} & \ms{E}{49.90}{0.58} \\
& \nolgo & Domain Best & \ms{A}{72.47}{3.26} & \ms{B}{27.44}{1.52} & \ms{C}{32.77}{0.68} & \ms{D}{21.75}{1.35} & \ms{E}{49.90}{0.58} \\
\cmidrule(l{2pt}){2-8}
& \cclogo CheetahClaw & Final & \pc{A}{54.83} & \pc{B}{26.39} & \pc{C}{32.94} & \pc{D}{18.62} & \pc{E}{49.23} \\
& \nolgo \texttt{ds-v4-flash} & Proxy Best & \pc{A}{54.42} & \pc{B}{25.51} & \pc{C}{29.22} & \pc{D}{18.12} & \pc{E}{48.77} \\
& \nolgo & Domain Best & \pc{A}{54.83} & \pc{B}{26.77} & \pc{C}{32.94} & \pc{D}{18.62} & \pc{E}{49.23} \\
\bottomrule
\end{tabular}}
\caption{\textbf{Main results of \benchmark on the generalization benchmarks.}
Each cell is the score followed by its change in points from the $M_0$ row of its block, the model before any curation.
\emph{Proxy Best} is the round whose average proxy score is the highest, and \emph{Domain Best} takes each domain from its own best round, so it is not a deployable checkpoint.
Proprietary systems are run with three seeds \citep{rank2026posttrainbench} and we report the mean.
}
\label{tab:main}
\end{table*}

\subsection{Research Agents}
We evaluate three research agents that pair a harness with a backbone model: Claude Code~\citep{anthropic2025claudecode} with \texttt{Opus-5}~\citep{anthropic2026opus5}, Codex~\citep{openai2025codexcli} with \texttt{GPT-5.6 Sol}~\citep{openai2026gpt56} (reasoning effort: medium), and CheetahClaw~\citep{cheetahclaws2026} with \texttt{deepseek-v4-flash}~\citep{xu2026deepseek}.
Each agent runs fully autonomously with all tool calls auto-approved, inside a sandbox whose network access is restricted to the HuggingFace Hub; the generalization benchmarks are never present in its environment.
All rounds of a run are carried out in a single agent session: at the start of each round the session is resumed with the new proxy results, so the agent retains its full interaction history from earlier rounds alongside the model pool, the data pool, and the written principles.

\subsection{Training Details}
Each round fine-tunes the checkpoint the agent selects with LoRA~\citep{hu2021lora} and AdamW~\citep{loshchilov2017decoupled}, for one epoch over the submitted pool.
The agent chooses the learning rate, LoRA rank, batch size, sequence length and warmup within a fixed range; the rest of the pipeline is frozen.
Each round is capped at $15$M training tokens and $40$k entries, and a run lasts five rounds.
Following PostTrainBench~\citep{rank2026posttrainbench}, we repeat every Claude Code and Codex configuration with three training seeds and report the mean across seeds; for open source systems such as CheetahClaw we run once per base model.

\begin{figure*}[!t]
  \centering
  \includegraphics[width=\textwidth]{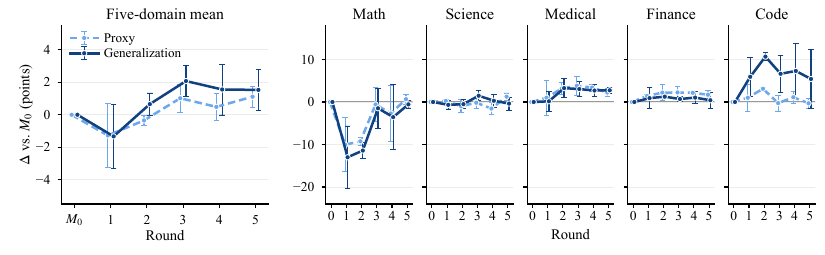}
  \caption{\textbf{Proxy gains generalize to held-out benchmarks.}
  We present the trend of the proxy gains and the generalization gains of model during training process on Qwen3-4B-Base with Codex and GPT-5.6 Sol.
  Proxy gains during training directly translate to generalization gains on the held-out benchmarks, indicating that the agent system is capable of curating post-training data for healthy post-training performance.}
  \label{fig:trend}
\end{figure*}

\subsection{Main Results}
We present the main results of the experiments in \cref{tab:main}.
For each agent run for each base model, we report three values for each domain.
\emph{Final} means the final model performance on the generalization benchmarks after the last round of training.
\emph{Proxy Best} is the checkpoint whose average proxy score is the highest, scored on the generalization benchmarks.
\emph{Domain best} means the best performance of the models in the loop on the generalization benchmarks for the corresponding domain.

\paragraph{Agent systems differ in how much they improve the model.}
Averaged over the five domains and the three base models, the proxy-best checkpoint gains $+2.3$ points over the starting model with Claude Code, $+0.6$ with Codex, and $-0.3$ with CheetahClaw.
In relative terms, computing $(s - s_{M_0})/s_{M_0}$ for each base model and domain and averaging the fifteen values with equal weight, the gains are $+9.2\%$, $+3.1\%$ and $+2.9\%$; the relative figure weights low-baseline cells heavily, so we report both.
The largest relative gain, $+63\%$ on medicine with Ministral-3-3B ($23.65 \rightarrow 38.46$), starts from a base model that often does not end its answer with an option letter, so part of this gain is the model learning the answer format rather than medical knowledge.
We find that the performance gain over different domains is not consistent.
The agents tend to improve the performance of the models on medical (4.3 absolute points) and code (3.2 absolute points) benchmarks, while the performance on math behaves differently as training downgrades the performance, and we conduct a detailed case analysis in \cref{app:case-math}.
We also observe that the different base models have different patterns of improvement.
The agents tend to improve more on Ministral-3-3B than on Qwen3-4B-Base and Olmo-3-1025-7B, indicating a high model-dependent effect of the current agent systems on autonomous post-training.

\paragraph{Later curation rounds can undo the gains of earlier ones.}
As the training process continues, the agent system may not maintain the performance gain over previous rounds on different domains simultaneously.
Across 5 domains over the 9 different agent-model setups, 24 out of 45 settings experience a performance drop comparing to the performance of the round with the highest proxy gain.
After gaining performance on the proxy benchmarks, what the agent system plans to do next is not guaranteed to be translated into further performance gain.
It may even undo the gains of previous rounds, suggesting that modern agentic systems have a price to pay for exploration: the checkpoint they hand back is often not the best one they have produced.
\Cref{app:case-pareto} follows one run in which a single round improves every domain at once and later rounds give part of that back.

\paragraph{No single round is best for all domains.}
The per-domain best row is an oracle: for each domain we take the highest accuracy on the generalization benchmarks any round reached, which allows a different checkpoint per domain and is therefore not a deliverable model.
Comparing it with the proxy-best row separates two costs, curating badly and having to hand back one model for five domains.
The second is almost universal: in $43$ of the $45$ model--domain cells the best round for a domain is not the round the agent selects, at a cost of $2.0$ points on average and up to $6.9$.
This suggests that the current agent systems do not have enough capability to promote the pareto frontier of the model capabilities which sustains already obtained capabilities while exploring new capabilities.

%% file: sections/4_analysis.tex
\section{Analysis}
\label{sec:analysis}

The main results in \cref{tab:main} report the endpoints of a run: the checkpoints an agent hands back after five rounds.
They leave open what happens inside a run, and which part of an agent system is responsible for the difference between systems.
We therefore ask three questions.
First, we follow a run round by round and ask whether the proxy scores the agent steers by move together with the generalization scores we report.
Second, we hold the backbone fixed and vary only the harness, which separates what the scaffold contributes from what the model driving it contributes.
Third, we look inside the training pools: which datasets each agent selects, how far different agents overlap, and when they stop searching for new sources.

\ifarxivbuild\else
\begin{figure}[t]
    \centering
    \includegraphics[width=\colfig]{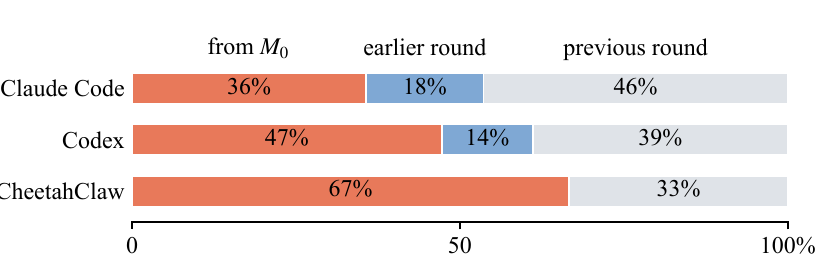}
    \caption{\textbf{The checkpoint each round is trained from.}
    Share of the rounds after the first that restart from $M_0$, branch from an earlier round, or continue from the round before.}
    \label{fig:restart}
\end{figure}
\fi

\ifarxivbuild\begin{figure*}[!b]\else\begin{figure*}[t]\fi
  \centering
  \begin{minipage}[b]{0.32\textwidth}\centering
    \includegraphics[width=\linewidth]{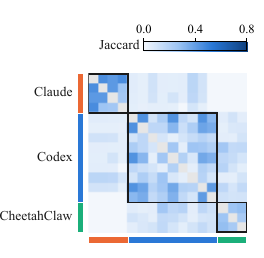}\\[-2pt]{\small (a)}
  \end{minipage}\hfill
  \begin{minipage}[b]{0.32\textwidth}\centering
    \includegraphics[width=\linewidth]{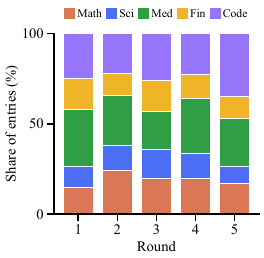}\\[-2pt]{\small (b)}
  \end{minipage}\hfill
  \begin{minipage}[b]{0.32\textwidth}\centering
    \includegraphics[width=\linewidth]{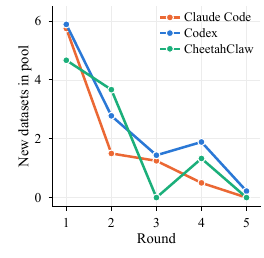}\\[-2pt]{\small (c)}
  \end{minipage}
  \caption{\textbf{What the agents train on.}
  (a) Jaccard overlap of the datasets used by each pair of runs; blocks group runs by agent.
  (b) Share of each round's training pool going to each domain, averaged over runs.
  (c) Datasets entering the training pool for the first time in each round, averaged over runs.}
  \label{fig:agent-data}
\end{figure*}

\subsection{Trend Analysis}
\ifarxivbuild
\begin{wrapfigure}{r}{0.46\linewidth}
  \vspace{-\intextsep}
  \centering
  \includegraphics[width=\linewidth]{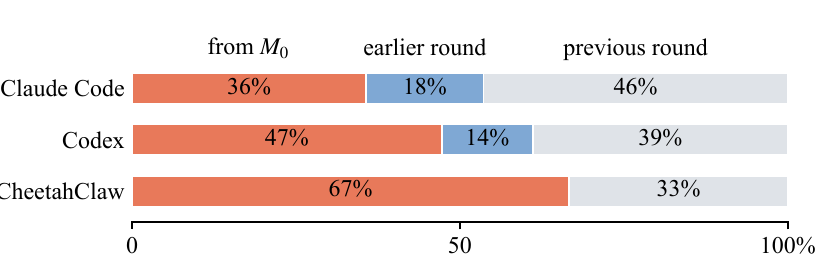}
  \caption{\textbf{The checkpoint each round is trained from.}
  Share of the rounds after the first that restart from $M_0$, branch from an earlier round, or continue from the round before.}
  \label{fig:restart}
\end{wrapfigure}
\fi
We present a study of trend analysis in \cref{fig:trend}.
We analyze with Qwen3-4B-Base as the base model for \benchmark training and Codex with GPT-5.6 Sol as the agent system.
We find that the proxy gains during training move in the same direction as the generalization gains on the generalization benchmarks.
Across different domains, the two signals do not move by the same amount.
On code the proxy gains stay within 3 points of the starting model, while the generalzation gains reach 11 points.
We also observe how the agent system handles the proxy benchmark feedback by selecting different checkpoints for each round in \cref{fig:restart}.
Claude Code and Codex tend to continue curating from an earlier checkpoint, in $64\%$ and $53\%$ of their rounds, while CheetahClaw starts from scratch in $67\%$ of its own rounds and never branches back to an older round at all.

\begin{takeaway}
\textbf{Takeaway.}
Proxy gains during training move in the same direction as the gains on the generalization benchmarks, and different agent systems have distinct restart strategies when encountering proxy benchmark feedback.
\end{takeaway}

\subsection{Harness Performance}

\ifarxivbuild
\begin{wrapfigure}{r}{0.46\linewidth}
  \vspace{-\intextsep}
  \centering
  \includegraphics[width=\linewidth]{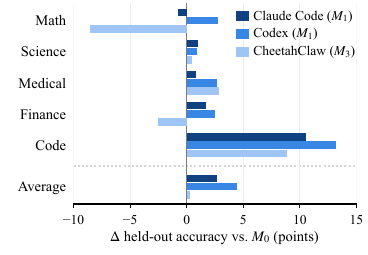}
  \caption{\textbf{Harnesses with a fixed backbone.} Change in generalization accuracy relative to $M_0$ at each harness's proxy-best checkpoint (in parentheses); DeepSeek V4.1 Flash Max on Qwen3-4B-Base, one run each.}
  \label{fig:harness}
\end{wrapfigure}
\else
\begin{figure}[t]
    \centering
    \includegraphics[width=\colfig]{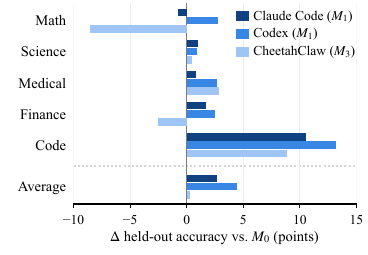}
    \caption{\textbf{Harnesses with a fixed backbone.} Change in generalization accuracy relative to $M_0$ at each harness's proxy-best checkpoint (in parentheses); DeepSeek V4.1 Flash Max on Qwen3-4B-Base, one run each.}
    \label{fig:harness}
\end{figure}
\fi

We analyze the performance of the same agent backbone with different harnesses to study the contribution of the harness to the autonomous post-training performance~\citep{gu2026model}.
To study this, we fix the backbone model as DeepSeek V4.1 Flash Max~\citep{xu2026deepseek} and vary the harness to be claude code, codex and cheetah claw.
We present the performance of the model at the proxy-best checkpoint for each harness in \cref{fig:harness}.
We observe that different harnesses produce different results.
On average, Codex produces the best results of 4.3 gains on the generalization benchmarks, followed by Claude Code with 2.6 points and CheetahClaw with 0.2 points.
On average, the agent systems contribute most significantly to the performance gain on coding domains, with the highest gains by Codex with 13.2 points.
This indicates that autonomous post-training performance should not only consider developing stronger models but also consider carefully crafting the harnesses to match model capabilities, so that better design of the harness can complement the strength of the model.

\begin{takeaway}
  \textbf{Takeaway.}
  Harness design is also crucial for autonomous post-training performance. With backbone fixed, the choice of harness contributes to the curation performance of the base model.
  \end{takeaway}

\subsection{Agent Decision-Making}
We further investigate the details of the agent decision-making process on dataset selection for post-training.
In \cref{fig:agent-data}, we present the datasets each agent selects into its training pool across runs and rounds.
We first observe that each agent has a preferred dataset taste.
Two runs of the same agent reach a Jaccard overlap of 0.24 on their dataset lists, against 0.07 for two runs of different agents, and the overlap does not rise when the two runs also share a base model (0.23 within a base model against 0.25 across base models).
This suggests that what an agent reaches for is set by the agent itself and carries over unchanged to a new target model, suggesting that current agent systems do not adapt to the target model.
Second, we observe that the share of training data of different domains is consistent.
Averaged over runs, the domain shares barely move from round to round. 
Individual runs are much less even: one Codex run devotes its entire final pool to code, and one CheetahClaw run gives medical over $80\%$ of two of its rounds.
Finally, fewer and fewer datasets enter the pool for the first time as the run goes on.
This suggests that current agent systems behave statically as the task horizon progresses: the agent is no longer willing to explore new sources of data to promote performance gains in later rounds.

\begin{takeaway}
\textbf{Takeaway.}
Different agent systems have their own dataset preferences, and exploration of new sources of data is not further promoted as the task horizon progresses.
\end{takeaway}

%% file: sections/5_related_work.tex
\section{Related Work}
\label{sec:related_work}

\subsection{Data Curation Benchmarks.}
Early data curation benchmarks score a curation pipeline by the downstream performance of the model it produces~\citep{mazumder2023dataperf, gadre2023datacomp, li2024datacomp}.
Recent work hands the pipeline to an agent: CurationBench~\citep{kang2026can} has it select from a fixed pool, PostTrainBench~\citep{rank2026posttrainbench} gives it a base model and a benchmark and lets it post-train against that benchmark, RSIBench-Data~\citep{meng2026rsibench} closes the loop on one benchmark, CurateEvo~\citep{wang2026curateevo} evolves the curation policy itself, BenchEvolver~\citep{wu2026benchevolver} evolves reference solutions into harder verifiable coding tasks for evaluation and training, and \citet{chi2026ai4ai} move the object of improvement from the data to the training algorithm.
\benchmark requires agents to focus on multiple domains simultaneously during the agentic data loop, allowing agents to select from dynamic data, and using generalization benchmarks to evaluate the genuine capability promotion.

\subsection{Recursive Self-Improvement Systems.}
Recursive self-improvement (RSI) dates back to the G\"odel machine~\citep{schmidhuber2007godel}, which rewrites its own code only under a proof of benefit.
LLM-based agents relax that proof into empirical validation, editing their own scaffold and keeping whichever variant scores higher~\citep{lu2024ai, zhang2026darwin}; later work makes the editing mechanism itself editable~\citep{zhang2026hyperagents}, shares experience across a population~\citep{weng2026group}, co-evolves the agent with its evaluator~\citep{iacob2026red}, turns the loop on algorithm discovery~\citep{novikov2025alphaevolve}, or lets several models improve one another by distilling their collaboration back into each model~\citep{feng2026single}.
Work on harnesses reports that a loop tuned on its own evolution split transfers only partly to tasks it never saw~\citep{xia2026rrsi, wu2026modularrsi, wang2026rethinking}, while a second route leaves the scaffold alone and updates the model's weights instead~\citep{hebbar2026sia}.
\benchmark instead fits into the RSI framework by benchmarking the capability of agent systems in diverse domain post-training data curation, which is critical for RSI systems to achieve better automation in real-world data curation.

%% file: sections/6_conclusion.tex
\section{Conclusion}
We present \benchmark, a benchmark for evaluating the capability of modern LLM-based AI agents in post-training curation of realistic tasks.
Instead of focusing on data curation to lift performance on a single benchmark, \benchmark aims to evaluate by focusing on curation of multiple domains reflecting the real-world post-training scenario.
To further mitigate the caveat of data contamination, \benchmark introduces proxy benchmarks and generalization benchmarks to avoid overfitting to the domains it seen during curation.
Across multiple experiments on \benchmark with different agent systems and base models, we observe that current agent systems can form a curation strategy from the feedback they see, yet lacking the capability of continuously improving the model, and often hand back suboptimal checkpoints.
Through further analysis, we find that the different agent systems have their own preferences in data selection, and different harnesses matter for successful post-training automation.
We hope \benchmark serves as a first step towards more realistic and comprehensive evaluation of post-training automation.

%% file: sections/7_limitation.tex
\section*{Limitations}
Our work can be extended in several ways.
First, we only experiment with three agent systems and three base models.
Future work can continue to explore different agent systems and models with larger scales, which could study the capability of current agent systems in broader scenarios.
Second, the three agent systems are moving targets. 
Our results are a snapshot of the harness and backbone versions available at the time of writing.
Third, we cover five domains and only English text. 
We leave the exploration of other languages and domains for future work.
We hope this work serves as a first step towards more realistic and comprehensive evaluation of post-training automation.

\section*{Ethics Statement}
We use only public datasets and published benchmarks, collect no user data, and run agents sandboxed with every entry screened against the evaluation items; nonetheless, agents choose their training data autonomously, inheriting the biases of the sources they reach for and adding a selection bias of their own; thus any real-world use should include bias audits, content filtering, and strict dataretention controls.

%% file: sections/appendix.tex
\section{The Curation Loop}
\label{app:loop}
The loop each run follows is stated below; \cref{sec:setup} describes it in prose.

\begin{algorithm}[t]
\caption{Multi-domain data curation in \benchmark.}
\label{alg:loop}
\begin{algorithmic}[1]
\Require initial model $M_0$; agent $\mathcal{A}$; training pipeline $\mathcal{S}$; proxy suite $\mathcal{T}_{\text{proxy}}$ over $N$ domains; generalization suite $\mathcal{T}_{\text{general}}$; per-round budget $\beta$; horizon $T$
\State $\mathcal{M}\gets\{M_0\}$,\; $P\gets\emptyset$,\; $\mathbf{s}_0\gets\mathcal{T}_{\text{proxy}}(M_0)$,\; $R_0\gets\mathcal{A}(\mathbf{s}_0)$
  \Comment{$\mathcal{M}$, $P$ and $R$ are the only state that crosses rounds}
\For{$t=1,\dots,T$}
  \State $P\gets P\cup\mathcal{A}.\textsc{Acquire}(R_{t-1})$
    \Comment{\textsc{Search}, \textsc{Select} and \textsc{Pull} on the Hub; entries keep their source and $P$ only grows}
  \State $M^{\text{par}},\,D_t,\,\mathcal{C}_t\gets\mathcal{A}.\textsc{Commit}(\mathcal{M},P,R_{t-1})$
    \Comment{parent checkpoint, training set and recipe, within $\beta$}
  \State $M_t\gets\mathcal{S}(M^{\text{par}},D_t;\mathcal{C}_t)$,\; $\mathcal{M}\gets\mathcal{M}\cup\{M_t\}$
  \State $\mathbf{s}_t,E_t\gets\mathcal{T}_{\text{proxy}}(M_t)$
    \Comment{per-domain scores and per-item results, both returned to the agent}
  \State $R_t\gets\mathcal{A}(R_{t-1},\mathbf{s}_{0:t},E_t)$
    \Comment{principles rewritten from the evidence, under the same $\beta$}
\EndFor
\State \Return $\mathcal{T}_{\text{general}}(M_T)$ and $\mathcal{T}_{\text{general}}(M_{t^{\star}})$, where $t^{\star}=\arg\max_t\operatorname{mean}(\mathbf{s}_t)$
  \Comment{scored once, after the run}
\end{algorithmic}
\end{algorithm}

\section{Prompts}
\label{app:prompts}
This section reproduces the text an agent sees in a run of \benchmark, excerpted verbatim from the round-5 workspace of one run; passages omitted here are marked [\ldots].
Only the round number, the model path and the counts the harness fills in at run time differ between rounds and runs.

\begin{promptbox}[promptorange]{Round brief (\texttt{BRIEF.md})}
\begin{Verbatim}
# ParetoBench -- round 5 of 5

You curate post-training data for `/path/to/models/Olmo-3-1025-7B`, round by round.
Each round you assemble a training pool; the harness trains on exactly that
pool, with the recipe you choose, **from a checkpoint you choose out of the model pool** (`"init_from"` in `train_config.json`; omitted = the newest checkpoint), and scores the result on
five proxy evaluations:

- **math** -- math500_proxy_200.jsonl
- **science** -- gpqa_main_200.jsonl
- **medical** -- medbullets_proxy_150.jsonl
- **finance** -- supergpqa_econ_proxy_200.jsonl
- **code** -- mbppplus_200.jsonl

[...]

Your objective is the whole vector: push the Pareto front of the five scores
outward relative to the base model. A gain on one axis bought with a loss on
another is recorded as exactly that.

These five proxies are your **feedback instrument, not the final exam**: the
run's reported result is measured afterwards on held-out sets you never see.
Not one item you are scored on here is in them. So what carries over to the
result is the capability; anything that fits only this proxy's surface does
not. Teaching the model to emit the answer format an axis happens to want will
move that axis's score here and buy nothing where the run is decided.
You may read the proxy questions and your model's own (wrong) answers --
`feedback/round_N/per_item/<axis>.jsonl` and `pb proxy peek --wrong` -- and
reason about where its ability falls short; that is the intended use.

[...]

Do not train on any evaluation set, its mirrors, or solutions to its items;
the checks below enforce this. What is forbidden and enforced: gold answers are never available; pages that
entries overlapping one are rejected at registration;
and searches containing proxy text are recorded as violations. **Overfitting
the proxies without real ability gains only hurts your held-out result.**

## The order of a round

0. Look at the data before you plan. What you already own is readable without
   limit: `pb candidates --domain <d> --in-pool 4 --preview-chars 400`
   (`--in-pool N` narrows to what round N actually trained on -- the store keeps
   every row ever registered, while a pool is the subset you chose), `--where
   "meta.subject==physics"` filters on the source's own columns, `pb candidates
   stats --by meta.<field>` profiles them, `pb show <id>` returns one row in full.
   One deadline only: **12 minutes into the round, the read routes refuse
   until `round_05/plan.json` exists** -- so look first and plan from
   what you saw. Once the plan is on disk, reading resumes; acquisition,
   registration, skill edits and `pb submit` are never gated by the clock, and
   the round's 30 minutes are yours to use after the plan too. `pb status`
   reports `plan.seconds_left_to_plan` and `seconds_left`.
1. Read `feedback/README.md` and what it points to. Read your skill at
   `.claude/skills/data-curation/SKILL.md`. In round 1 the skill is empty: fill
   every required slot with what you believe, tagged `prior`, before anything else.
2. Write `round_05/plan.json`: the steps you will take this round,
   each tied to principles in your skill --
   `[{"action": "...", "because": ["<principle id>", ...], "expect": {"<axis>": "+2 to +4"}, "budget_usd": ...}, ...]`.
   A principle id is the `[id | basis]` tag of a bullet in your skill (e.g.
   `math1`), not a section title. Acquisition commands refuse to run until the
   plan exists and every cited id is an active principle. The plan is the
   round's audit record; if next round's you should remember what you expected
   and why, write that into the skill -- feedback carries scores, not plans.
3. Acquire and register. Decide what to select. Write `round_05/selection.jsonl`
   and `train_config.json`.
4. Revise the skill: cite evidence (`rN:axis+points`, checked against the scores),
   retire what was falsified. Run `pb skill check` after editing -- a skill that
   fails validation at round end is silently reverted to the issued version,
   and new bullets belong under the existing section headings, never new ones.
5. `pb submit round_05/selection.jsonl --train-config round_05/train_config.json`, then exit.

## How data comes in (through `pb`; no keys in your environment)

| source | commands | cost | provenance the gateway records |
|---|---|---|---|
| **Select** from Hugging Face | no seeds in this run: there is no starting point and finding sources is yours to do -- `pb hf schema` + `pb hf pull --instruction-field/--output-field` on any repo (search with `pb hf search`). Evaluation sources are refused at pull, with the reason | bytes: 20 GB per run | repo, split, row |

That is the only way in. Native `WebSearch` is still yours for DISCOVERY --
finding which repos exist is most of the problem on this track -- but a page's
text is not a way into the store: everything trained on is rows pulled from a
Hugging Face dataset, or your own `derived`/`author` rows built from them.

Pulled rows never enter your workspace, and they do not need registering: a
pull puts them straight into the candidate store and replies with a preview,
the distribution, and **the ids**, plus anything decontamination or the
duplicate check refused. Filter at pull time with `--where`, or later at
selection time with `pb candidates --where`. Text you wrote or transformed
yourself is the one thing that registers from a file (`pb register
entries.jsonl`): `derived` must name the stored row it came from, `author` is
counted separately -- see CHANNELS.md. `pb show` returns one full row; reading
what you already own is counted but not capped.

## The data pool

- **Candidate store** (persists across rounds): everything registered, with a
  harness-assigned id. **The candidate store already holds 91,590 registered entries** from earlier rounds. Look at them before pulling more: `feedback/candidates.json`, `pb candidates stats --by meta.source`, `pb candidates --domain <d>`. They are selectable as they are; nothing has to be re-downloaded. The store is what this run has pulled, not what exists: the repos pulled from so far hold 3,067,241 rows upstream, of which 107,200 have been pulled (`feedback/candidates.json` itemises them; `pb seed list` names reference repos this run has not touched). A filter that empties the store has not emptied the source -- pull again with a new `--offset` or `--where`.
- **This round's pool P_t** = the ids you list in `selection.jsonl`
  (`{"id","reason"}` per line; a reason is required on newly added entries).
  The harness pulls the text itself.
- **What enters**: only entries that pass 13-gram decontamination against every
  proxy and held-out item, each with a provenance the gateway wrote. Rejections
  come back with reasons.
- **How much**: `sum(min(len, max_len)) <= 15,000,000` tokens per round, over a
  pool of at most 40,000 entries. **Training is one pass: `epochs` is fixed at 1**,
  so the token budget is a budget for DATA, not for repetition -- filling it
  means about 34,800 entries of mixed length, more if they are short. That is
  several times what one `pull` returns, so plan on many pulls across the
  round. **Aim to use the full budget**: tokens you do not schedule are simply
  lost. Check the estimate `pb submit` would see (`pb status` /
  `feedback/budget.json`) as you go, and stop short of the cap only for a
  reason you can state. Recipe keys you may set in `train_config.json`: `lr`, `lora_rank`, `batch_size`, `max_len`, `warmup_steps` (pinned: `epochs`=1);
  the seed is fixed too.
  **The model pool**: every scored round that saved a state, plus `base`.
  `pb status` lists it under `model_pool` with each checkpoint's parent,
  recipe and scores; name your choice with `"init_from": "<name>"` in
  `train_config.json` (`base`, or any `round_NN`; omitted = the newest
  checkpoint). `lora_rank` must match the chosen checkpoint's rank -- only
  `base` accepts any rank.
  Training is cumulative ALONG THE LINEAGE YOU PICK: what an ancestor
  learned -- or damaged -- is inherited by every descendant, and there is no
  undo other than branching from an earlier checkpoint (or `base`). An entry
  is **kept** if it lies on the chosen checkpoint's lineage (this pass is a
  later epoch for it) and **new** otherwise; `commit.json` reports the split
  relative to your chosen parent. A checkpoint that a later round regressed
  from is still in the pool -- branching from it discards the regression at
  the price of also discarding whatever else that round taught.
- **What it is used for**: one LoRA SFT pass over P_t, from the pool checkpoint you name in train_config.json (newest by default),
  then the five proxy evaluations. Nothing else. **Training is cumulative along the chosen lineage**: a bad round is escaped only by branching from an earlier checkpoint, at the price of re-teaching whatever later rounds added. per_item shows which items the CHOSEN parent gets wrong -- read it for the checkpoint you branch from, not just the latest.

## Budget this round

About 30 minutes and $3 of your own
model budget; there is no automatic submission extension. Reserve the final 5 minutes
for selecting existing candidates, writing train_config.json, updating the skill and
submitting. Remote acquisition closes at that point. Training and evaluation are
paid by the harness -- the compute per round is fixed, so they are not yours to
economise on. `pb status` reports time left and every balance.
\end{Verbatim}
\end{promptbox}

\begin{promptbox}[promptblue]{Acquisition channels (\texttt{CHANNELS.md})}
\begin{Verbatim}
# Channels and provenance

Every candidate enters the store with a provenance the harness wrote, then
passes 13-gram decontamination against every proxy and held-out item.
Contaminated rows are rejected outright and listed in the reply.

| channel | how a row gets in | provenance |
|---|---|---|
| `hf` | `pb hf pull` registers every mapped row on arrival | repo, split, row |
| `derived` | your `pb register entries.jsonl` citing a stored parent | parent_id, transform |

Rewrote a stored row?  Register the rewrite as `derived` with that row's id as
parent.

`output` is what the model will be trained to produce.

Blocked at the source: the HF repos that publish the proxy evaluations
themselves.  Everything else is decided by decontamination.
\end{Verbatim}
\end{promptbox}

\begin{promptbox}[promptgreen]{Feedback directory (\texttt{feedback/README.md})}
\begin{Verbatim}
Built by the harness at init and after every scored round into

    <run>/workspace/feedback/

and regenerated from the run's own records each time, so the agent never
needs a command to learn where it stands.  Harness-owned; the agent edits only
its skill and its round scratch directory.

    feedback/README.md              what each file is
    feedback/scores.tsv             round x axis, with M_0; non-dominated
    feedback/scores.json            the full history: vectors, Pareto, pool mix, recipe
    feedback/budget.json            what is left: rounds, compute cap, bytes, views
    feedback/candidates.json        what is registered: by domain / channel / source, lengths;
                                    `upstream` = rows still in each source vs rows pulled,
                                    for the repos this run pulled from (untouched
                                    reference seeds live in `pb seed list`)
    feedback/round_NN/
        eval.json                   per axis: score [...], delta vs M_0 and vs previous
                                    round, truncation rate, mean output tokens, n. [...]
        per_item/<axis>.jsonl       one scored item per line, {id, correct, solve_rate,
                                    model_answer, ...the question and its metadata}; gold
                                    never included. [...]
                                    200 lines, each a few thousand characters: slicing it
                                    with `head -c` shows one or two, so select with jq or
                                    grep (e.g. only the lines where correct is false)
        train.json                  loss first/last/min, steps, tokens, wall clock, recipe
        commit.json                 what changed in the pool vs what the chosen
                                    parent checkpoint had already trained on
        pool.json                   composition of P_t: by domain / channel / source, token quantiles

Gold answers are never written here; neither are other rounds' per-entry
reasons (the skill is the only channel for that).  Training and evaluation
dollars are not here either: the per-round compute is fixed, so they are the
harness's bill, not a variable the agent controls.
\end{Verbatim}
\end{promptbox}

\begin{promptbox}[promptpurple]{Skill template issued in round~1 (\texttt{SKILL.md})}
\begin{Verbatim}
---
name: data-curation
description: Accumulated principles for allocating a training-entry budget across
  math, science, medical, finance, code. Read this at the start of every round, before deciding anything, and
  revise it at the end of the round.
---

<!-- pb:harness:begin baseline -- written by the harness, do not edit -->
## Baseline and history

| axis | n | M_0 |
|---|---|---|
| math | full | 0.749 |
| science | full | 0.250 |
| medical | full | 0.340 |
| finance | full | 0.190 |
| code | full | 0.438 |

<!-- pb:harness:end baseline -->

## How to maintain this file

Your scratch files are discarded when the round ends -- `round_NN/` is archived
away -- but your own conversation is not: one session spans the whole run, so
you keep what you have already reasoned about. Past the context window that
becomes whatever auto-compaction leaves, which is not yours to choose. This
file is what you deliberately keep: write here what must survive that.

It is a template: every slot below is a dimension of the decision you make
each round.  Fill each required slot with at least one bullet -- in round 1
that will be a belief you hold before any evidence, tagged `prior`.  The slot
names say what to have an opinion about; they say nothing about what that
opinion should be.

Every bullet is tagged `- [id | basis] text`:

- `prior` -- you believe it, but no round has demonstrated it yet.
- `r2:finance+6.3,medical-2.1` -- round 2 moved those axes by those points.
  The harness checks these numbers against the recorded scores, so cite them
  accurately or the round is rejected.
- `retired:r4` -- you gave up on it in round 4. Move the bullet to `## Retired`
  and say in the text what falsified it.

The budget is 8000 tokens and it counts only the active bullets, so retiring
something you no longer believe gives you room to write something you do.

## Per-domain

### math
<!-- What data moves this axis, what does not, and where its ceiling is. -->

### science
<!-- What data moves this axis, what does not, and where its ceiling is. -->

### medical
<!-- What data moves this axis, what does not, and where its ceiling is. -->

### finance
<!-- What data moves this axis, what does not, and where its ceiling is. -->

### code
<!-- What data moves this axis, what does not, and where its ceiling is. -->

## Per-channel
<!-- What each source is worth and what it costs: which repos, which filters, which offsets. -->

## Recipe
<!-- What you have learned about lr / epochs / max_len / rank / batch. -->

## Interference
<!-- Cross-domain effects: data admitted for one axis moving another. -->

## Evaluation (optional)
<!-- Observations about the proxy evaluations themselves (noise, coverage, quirks). -->

## Allocation (optional)
<!-- Your standing allocation intent and why (one or two bullets). -->

## Retired

\end{Verbatim}
\end{promptbox}

\section{A Round in Practice}
\label{app:typical-round}
We present a typical round of how a agent curates a dataset in practice.
Below is a transcript of a round of Codex with \texttt{GPT-5.6 Sol} on Olmo-3-1025-7B, round 2.
This reflects the typical workflow of a agent curating a dataset in practice.

\begin{roundbox}{One round: Codex with \texttt{GPT-5.6 Sol} on Olmo-3-1025-7B, round 2}
\stepline{1}{Read the feedback}{The harness writes the five proxy scores and the per-item results into \texttt{feedback/}, and pastes the change from round~1 into the top of the notes file: math $-23.8$, code $+10.4$, medical $-4.7$, finance $-2.5$, science $0.0$.}
\stepline{2}{Read its own notes}{Nine beliefs carried over from round~1, each tagged with the round and the score change it rests on.}
\stepline{3}{Write a plan}{Four actions, each naming the beliefs it follows and the points it expects to move. The planning window is twelve minutes and the agent uses all of it.}
\stepline{4}{Look for sources}{One web search and one direct query to the Hugging Face API for a repository's file layout.}
\stepline{5}{Pull rows}{Four pulls, $17{,}100$ rows, each with an explicit field mapping such as \texttt{Question->Response}. Registration runs 13-gram decontamination against every proxy and generalization item and returns whatever it refuses.}
\stepline{6}{Choose the pool}{$25{,}908$ entries and $10.0$M scheduled tokens against a cap of $15$M. The agent opens no row: \texttt{full\_row\_views} is $0$.}
\stepline{7}{Choose the recipe and the parent}{Learning rate $2\times10^{-6}$, rank $32$, batch $64$, length $4096$, and \texttt{init\_from: base}, which discards the round-1 checkpoint.}
\stepline{8}{Train, evaluate, rewrite the notes}{Ten new beliefs and two retired. The harness checks the score change each belief cites against the record and rejects the round if they disagree.}
\end{roundbox}

Two steps carry most of what the later sections examine.
Step~7 is where an agent decides whether a run accumulates or restarts, and the agents differ sharply on it.
Step~6 is where the pool is fixed, and the zero in it holds for $49$ of the $50$ rounds we logged: an agent settles the fate of tens of thousands of entries without reading one of them.

\section{Case Studies}
\label{app:case-studies}
\subsection{Failure Case on Mathematical Reasoning}
\label{app:case-math}

We observe that in \cref{tab:main} that most of the agents fail to improve the capability of the original model on mathematical reasoning tasks.
In this section we provide a detailed case study of this to understand the failure mode of the agents.
We study the run of Claude Code with Opus-5 on Olmo-3-1025-7B, where math falls by $20.1$ points in the first round and recovers above $M_0$ in the second, with both rounds trained from the base model under the same recipe.
We observe that the agents tend to select datasets with shorter answer lengths such as NuminaMath~\citep{numina_math_datasets} in the earliest rounds, where the original model tends to generate longer answers.
Such behavior is detrimental to the improvement of the model on mathematical reasoning tasks, since the model gains accuracy by generating longer answers.
In later rounds, the agents tend to change its strategy of dataset selection to select datasets with answers of similar length to the original model, usually datasets with longer reasoning chains~\citep{moshkov2025aimo2}.
The selection record from the agent also shows the incentive at work: in the first round it filters NuminaMath~\citep{numina_math_datasets} datasets for ``concise math CoT ($\le$5k chars) to finish solutions within budget'', since a token-denominated budget buys more entries when the answers are short. In the second round it discards that lineage and pulls long \texttt{<think>} traces instead, recovering math with a seventh of the entries.

\begin{figure}[t]
    \centering
    \includegraphics[width=\colfig]{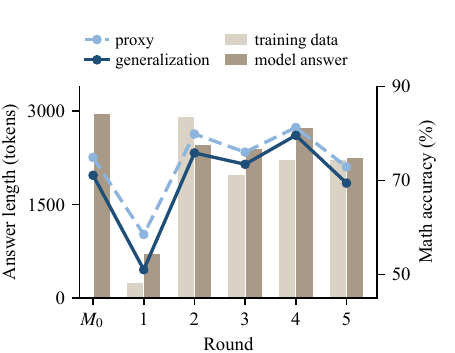}
    \caption{\textbf{Answer length and math accuracy over the rounds.}}
    \label{fig:case-math}
  \end{figure}

\subsection{Forming Experience While Curating}
\label{app:case-experience}
In the execution of \benchmark, the agent forms experience while curating the datasets by weaving its own observations into the notes file.
Such note files document how agent forms its own beliefs with prior experiences and observations on the proxy benchmark.
In the box below, we present a case study of the agent's experience formation while curating the finance domain in the run of Claude Code with Opus-5 on Olmo-3-1025-7B.

\begin{notebox}{Curation notes, \texttt{finance} slot, Claude Code with Opus-5 on Olmo-3-1025-7B}
\noteline{fin1}{prior}{SuperGPQA econ at 0.19, 116/200 [\ldots] (10-option MCQ, 27\% truncation): biggest headroom; economics/finance reasoning QA (WebInstruct econ, finance instruct sets, 10-option science-style MCQ reasoning) is the lever.}

\roundline{Round 1}{a short-answer pool; every domain but finance loses its reasoning style.}%
{finance $0.188 \rightarrow 0.224$. Answers stay at $2{,}279$ tokens, against $65$--$718$ on the other four domains.}

\noteline{fin2}{r1:finance+4.0}{The only r1 axis that kept <think> reasoning (2279 tok) matched the 10-option OpenScience econ traces; 17 [\ldots] items reached >=0.5. More 10-option econ/business MCQ reasoning traces is the lever.}

\roundline{Round 2}{acts on \texttt{fin2} and makes 10-option econ traces the finance data.}%
{finance $0.244$. Truncation rises to $29\%$ at $3{,}356$ tokens.}

\noteline{fin3}{r2:finance+3.5}{Second confirmation for fin2: 10-opt OpenScience econ traces again lifted finance (mean solve 0.188->0.244, 65 up/38 down, 18 [\ldots] items to >=0.5), but truncation is 29\% at 3.4k tok -{}- shorter traces should help further.}

\roundline{Round 3}{acts on \texttt{fin3} and filters the same source to traces of at most 8k characters.}%
{finance $0.260$, the best in the run. Truncation falls to $21\%$ at $2{,}726$ tokens.}

\noteline{fin4}{r3:finance+2.5}{Third straight finance gain from 10-opt econ traces (0.225->0.25; mean solve 0.244->0.26); truncation fell 29\%->21\% with <=8k-char traces. Keep feeding fresh-offset econ traces.}
\end{notebox}

We observe that the agent forms its own useful experience and leads to continuous improvement during later rounds.
In this case, after first round of curation, the agent forms a belief that the 10-option OpenScience~\citep{nvidia2025openscience} econ traces are the most effective for the finance domain after observing that finance is the only domain whose answers keep their long reasoning style after round~1, at $2{,}279$ tokens against $65$--$718$ elsewhere.
In the third round, the agent further forms a belief that shorter traces should help, and acts on it in the round that follows.
This experience helps the agent to further improve the finance accuracy to $0.250$, the best in the run, suggesting that the agent's experience formation is effective in guiding its curation strategy.

\subsection{A Round That Improves Every Domain}
\label{app:case-pareto}
\begin{figure}[t]
    \centering
    \includegraphics[width=\colfig]{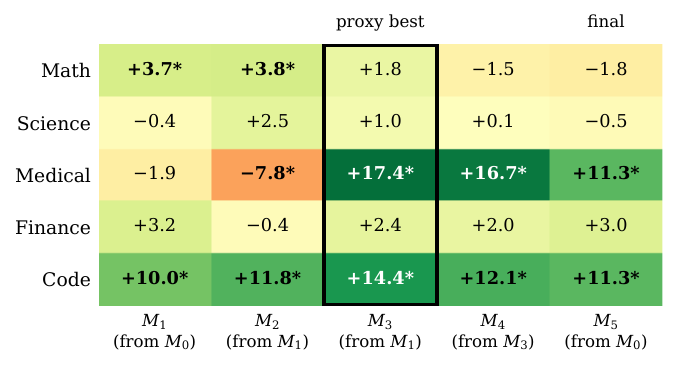}
    \caption{\textbf{Change on the generalization benchmarks relative to $M_0$, in points,} for Claude Code with \texttt{Opus-5} on Ministral-3-3B.
    Each column is one round's checkpoint and the checkpoint it was trained from.
    $*$ marks a change whose 95\% paired bootstrap interval over items excludes zero.
    The boxed round is at least as good as $M_0$ in every domain.}
    \label{fig:case-pareto}
\end{figure}

We provide a case from our runs in which the agent pushes the Pareto frontier forward: \cref{fig:case-pareto} follows Claude Code with \texttt{Opus-5} on Ministral-3-3B.
Round~2 continues from round~1 and trades medicine for math and code: it gains on both but loses $7.8$ points on medicine.
The agent then discards round~2: round~3 branches back from round~1 and rebuilds its medical data, and the resulting checkpoint is at least as good as $M_0$ on every generalization benchmark, significantly better on medicine and code, and also the round with the highest proxy score.
The improvement does not persist: rounds~4 and~5 keep the medical and code gains but give back math, so the final checkpoint no longer dominates $M_0$.
The loop can thus reach a Pareto improvement over the starting model, and holding on to it is harder than finding it.

\section{Datasets the Agents Select}
\label{app:datasets}
\Cref{tab:data-claude,tab:data-codex,tab:data-cheetah} list the datasets each agent selects most often: Claude Code reaches for newer and more varied sources, while Codex and CheetahClaw stay with older ones.

\begin{table*}[t]
\centering
\small
\setlength{\tabcolsep}{6pt}
\begin{tabular}{@{}l l@{}}
\toprule
\textbf{Domain} & \textbf{Dataset} \\
\midrule
\multirow{4}{*}{Math} & \texttt{nvidia/OpenMathInstruct-2} \citep{toshniwal2024openmath2} \\  
 & \texttt{AI-MO/NuminaMath-CoT} \citep{numina_math_datasets} \\  
 & \texttt{open-r1/Mixture-of-Thoughts} \citep{openr1} \\  
 & \texttt{nvidia/OpenMathReasoning} \citep{moshkov2025aimo2} \\  
\midrule
\multirow{5}{*}{Science} & \texttt{TIGER-Lab/WebInstructSub} \citep{yue2024mammoth2} \\  
 & \texttt{nvidia/OpenScience} \citep{nvidia2025openscience} \\  
 & \texttt{MegaScience/MegaScience} \citep{fan2025megascience} \\  
 & \texttt{allenai/tulu-3-sft-olmo-2-mixture} \citep{lambert2024tulu} \\  
 & \texttt{open-r1/Mixture-of-Thoughts} \citep{openr1} \\  
\midrule
\multirow{5}{*}{Medical} & \texttt{HPAI-BSC/medmcqa-cot-llama31} \citep{gururajan2024aloe, pmlr-v174-pal22a} \\  
 & \texttt{HPAI-BSC/MedQA-Mixtral-CoT} \citep{gururajan2024aloe, jin2020disease} \\  
 & \texttt{FreedomIntelligence/medical-o1-reasoning-SFT} \citep{chen2024huatuogpto1medicalcomplexreasoning} \\  
 & \texttt{nvidia/OpenScience} \citep{nvidia2025openscience} \\  
 & \texttt{glaiveai/reasoning-v1-20m} \citep{glaiveai2025reasoningv1} \\  
\midrule
\multirow{5}{*}{Finance} & \texttt{Josephgflowers/Finance-Instruct-500k} \citep{josephgflowers2025financeinstruct} \\  
 & \texttt{Jackrong/financial-economics-reasoning} \citep{shi2024industryinstruction} \\  
 & \texttt{teknium/dataforge-economics} \citep{teknium2023dataforgeeconomics} \\  
 & \texttt{nvidia/OpenScience} \citep{nvidia2025openscience} \\  
\midrule
\multirow{1}{*}{Code} & \texttt{ise-uiuc/Magicoder-OSS-Instruct-75K} \citep{wei2024magicoder} \\  
\bottomrule
\end{tabular}
\caption{\textbf{Datasets Claude Code selects most often.}}
\label{tab:data-claude}
\end{table*}

\begin{table*}[t]
\centering
\small
\setlength{\tabcolsep}{6pt}
\begin{tabular}{@{}l l@{}}
\toprule
\textbf{Domain} & \textbf{Dataset} \\
\midrule
\multirow{5}{*}{Math} & \texttt{nvidia/OpenMathInstruct-1} \citep{toshniwal2024openmath} \\  
 & \texttt{meta-math/MetaMathQA-40K} \citep{yu2023metamath} \\  
 & \texttt{TIGER-Lab/MathInstruct} \citep{yue2023mammoth} \\  
 & \texttt{NuminaMath-1.5} \citep{numina_math_datasets} \\  
\midrule
\multirow{5}{*}{Science} & \texttt{dkoterwa/camel\_ai\_chemistry\_instruction\_dataset} \citep{li2023camel} \\  
 & \texttt{dkoterwa/camel\_ai\_physics\_instruction\_dataset} \citep{li2023camel} \\  
 & \texttt{dkoterwa/camel\_ai\_biology\_instruction\_dataset} \citep{li2023camel} \\  
 & \texttt{zd21/SciInstruct} \citep{zhang2024sciinstruct} \\  
 & \texttt{allenai/sciq} \citep{SciQ} \\  
\midrule
\multirow{5}{*}{Medical} & \texttt{FreedomIntelligence/medical-o1-reasoning-SFT} \citep{chen2024huatuogpto1medicalcomplexreasoning} \\  
 & \texttt{medalpaca/medical\_meadow\_wikidoc} \citep{han2023medalpaca} \\  
 & \texttt{medalpaca/medical\_meadow\_medical\_flashcards} \citep{han2023medalpaca} \\  
 & \texttt{medalpaca/medical\_meadow\_medqa} \citep{han2023medalpaca} \\  
 & \texttt{keivalya/MedQuad-MedicalQnADataset} \citep{ben2019question} \\  
\midrule
\multirow{5}{*}{Finance} & \texttt{gbharti/finance-alpaca} \citep{alpaca} \\  
 & \texttt{Jackrong/financial-economics-reasoning} \citep{shi2024industryinstruction} \\  
 & \texttt{gbharti/finance-alpaca-csv} \citep{alpaca} \\  
 & \texttt{sujet-ai/Sujet-Finance-Instruct-177k} \citep{sujet2024financeinstruct} \\  
\midrule
\multirow{5}{*}{Code} & \texttt{iamtarun/python\_code\_instructions\_18k\_alpaca} \citep{iamtarun2023python18k} \\  
 & \texttt{mlabonne/Evol-Instruct-Python-26k} \citep{zeng2024automatic} \\  
 & \texttt{ise-uiuc/Magicoder-Evol-Instruct-110K} \citep{wei2024magicoder} \\  
 & \texttt{ise-uiuc/Magicoder-OSS-Instruct-75K} \citep{wei2024magicoder} \\  
\bottomrule
\end{tabular}
\caption{\textbf{Datasets Codex selects most often.}}
\label{tab:data-codex}
\end{table*}

\begin{table*}[t]
\centering
\small
\setlength{\tabcolsep}{6pt}
\begin{tabular}{@{}l l@{}}
\toprule
\textbf{Domain} & \textbf{Dataset} \\
\midrule
\multirow{3}{*}{Math} & \texttt{openai/gsm8k} \citep{cobbe2021gsm8k} \\  
 & \texttt{mlfoundations-dev/numina-math-olympiads-judged} \citep{numina_math_datasets} \\  
 & \texttt{AI-MO/NuminaMath-1.5} \citep{numina_math_datasets} \\  
\midrule
\multirow{5}{*}{Science} & \texttt{avaliev/ChemistryQA} \citep{wei2020chemistryqa} \\  
 & \texttt{mlfoundations-dev/camel-ai-biology} \citep{li2023camel} \\  
 & \texttt{mlfoundations-dev/camel-ai-chemistry} \citep{li2023camel} \\  
 & \texttt{mlfoundations-dev/camel-ai-physics} \citep{li2023camel} \\  
 & \texttt{allenai/sciq} \citep{SciQ} \\  
\midrule
\multirow{3}{*}{Medical} & \texttt{medalpaca/medical\_meadow\_medqa} \citep{han2023medalpaca} \\  
 & \texttt{medalpaca/medical\_meadow\_medical\_flashcards} \citep{han2023medalpaca} \\  
 & \texttt{medalpaca/medical\_meadow\_wikidoc} \citep{han2023medalpaca} \\  
\midrule
\multirow{2}{*}{Finance} & \texttt{gbharti/finance-alpaca} \citep{alpaca} \\  
 & \texttt{ssbuild/alpaca\_finance\_en} \citep{alpaca} \\  
\midrule
\multirow{2}{*}{Code} & \texttt{sahil2801/CodeAlpaca-20k} \citep{alpaca} \\  
 & \texttt{iamtarun/python\_code\_instructions\_18k\_alpaca} \citep{iamtarun2023python18k} \\  
\bottomrule
\end{tabular}
\caption{\textbf{Datasets CheetahClaw selects most often.} }
\label{tab:data-cheetah}
\end{table*}

\clearpage
\section{Human Baseline}
\label{app:human-baseline}
We compare the agents with human supervision within the curation loop of \benchmark (\cref{alg:loop}).
The agent searches for and pulls candidate datasets from the Hugging Face Hub as in a regular run, and a human then takes over the commit step: the human reviews the entries the agent pulled and decides which of them enter the training pool, under the same token budget, training pipeline and evaluation as the agents.
\Cref{tab:human-baseline} reports the result on the three base models beside $M_0$ and the strongest agent.

\ifarxivbuild\begin{table}[!ht]\else\begin{table*}[t]\fi
\centering
\small
\setlength{\tabcolsep}{4pt}
\begin{tabular}{@{}l l ccccc@{}}
\toprule
\textbf{Base Model} & \textbf{Model} & \textbf{Math} & \textbf{Science} & \textbf{Medical} & \textbf{Finance} & \textbf{Code} \\
\midrule
\multirow{3}{*}{Qwen3-4B-Base} & $M_0$ & 65.33 & 31.94 & 39.36 & 24.50 & 65.03 \\
 & Claude Code + \texttt{Opus-5}, Proxy Best & 60.97 & 33.33 & 43.69 & 24.50 & 71.93 \\
 & Human supervised & 72.50 & 25.88 & 40.37 & 28.00 & 65.95 \\
\midrule
\multirow{3}{*}{Ministral-3-3B} & $M_0$ & 38.92 & 27.65 & 23.65 & 17.62 & 26.69 \\
 & Claude Code + \texttt{Opus-5}, Proxy Best & 34.64 & 29.00 & 38.46 & 19.63 & 39.26 \\
 & Human supervised & 65.67 & 19.70 & 39.70 & 20.00 & 50.46 \\
\midrule
\multirow{3}{*}{Olmo-3-1025-7B} & $M_0$ & 71.08 & 22.73 & 32.43 & 18.25 & 48.01 \\
 & Codex + \texttt{GPT-5.6 Sol}, Proxy Best & 69.50 & 24.92 & 30.41 & 19.00 & 49.90 \\
 & Human supervised & 68.00 & 24.37 & 32.94 & 24.25 & 48.93 \\
\bottomrule
\end{tabular}
\caption{\textbf{Human baseline on the generalization benchmarks.} $M_0$ and agent rows are taken from \cref{tab:main}; the agent row is the agent whose \emph{Proxy Best} checkpoint has the highest mean on that base model.}
\label{tab:human-baseline}
\ifarxivbuild\end{table}\else\end{table*}\fi

Human in the loop remains crucial for current agent systems to improve a model across domains.
Human supervision reaches gains that no agent reaches, most clearly on math for Qwen3-4B-Base and Ministral-3-3B and on finance for Olmo-3-1025-7B, and on average it matches or exceeds the strongest agent on every base model.
This motivates future RSI systems to keep humans in the curation loop and learn from their decisions, rather than curating fully autonomously.

\section{Variance Across Seeds}
\label{app:seed-variance}
\Cref{tab:main} reports the mean over three training seeds for Claude Code and Codex; \cref{tab:seed-variance} adds the sample standard deviation across seeds on Qwen3-4B-Base.
Most standard deviations lie between one and three points; the largest are on math for Claude Code ($\pm 5.6$ at Final) and on code for Codex ($\pm 6.9$).

\ifarxivbuild\begin{table}[!ht]\else\begin{table*}[t]\fi
\centering
\small
\setlength{\tabcolsep}{4pt}
\begin{tabular}{@{}l l ccccc@{}}
\toprule
\textbf{Agent} & \textbf{Checkpoint} & \textbf{Math} & \textbf{Science} & \textbf{Medical} & \textbf{Finance} & \textbf{Code} \\
\midrule
$M_0$ & -- & 65.33 & 31.94 & 39.36 & 24.50 & 65.03 \\
\midrule
Claude Code & Final & 58.67{\scriptsize$\pm$5.62} & 30.13{\scriptsize$\pm$1.05} & 40.76{\scriptsize$\pm$1.03} & 22.67{\scriptsize$\pm$2.84} & 70.50{\scriptsize$\pm$1.68} \\
\texttt{Opus-5} & Proxy Best & 60.97{\scriptsize$\pm$1.70} & 33.33{\scriptsize$\pm$2.67} & 43.69{\scriptsize$\pm$2.73} & 24.50{\scriptsize$\pm$0.87} & 71.93{\scriptsize$\pm$1.96} \\
\midrule
Codex & Final & 65.11{\scriptsize$\pm$0.85} & 31.57{\scriptsize$\pm$1.53} & 42.12{\scriptsize$\pm$0.71} & 24.91{\scriptsize$\pm$1.83} & 63.86{\scriptsize$\pm$6.88} \\
\texttt{GPT-5.6 Sol} & Proxy Best & 65.11{\scriptsize$\pm$0.85} & 31.57{\scriptsize$\pm$1.53} & 42.12{\scriptsize$\pm$0.71} & 24.91{\scriptsize$\pm$1.83} & 63.86{\scriptsize$\pm$6.88} \\
\bottomrule
\end{tabular}
\caption{\textbf{Mean and sample standard deviation over three training seeds on Qwen3-4B-Base}, on the generalization benchmarks.}
\label{tab:seed-variance}
\ifarxivbuild\end{table}\else\end{table*}\fi

